\documentclass[a4paper]{article}

\usepackage{INTERSPEECH2020}
\usepackage{comment}
\usepackage{multirow}
\usepackage{colortbl}
\usepackage{balance}
\usepackage[utf8x]{inputenc}

\title{Exploiting Spectral Augmentation for Code-Switched Spoken Language Identification}
\name{Pradeep Rangan$^*$, Sundeep Teki$^*$, and Hemant Misra}
\address{
  Applied Research Department, Swiggy (Bundl Technologies India Limited)
}
\email{[pradeep.r1, sundeep.teki, hemant.misra] @swiggy.in} 
\begin{document}

\maketitle

\def\thefootnote{*}\footnotetext{These authors contributed equally to this work}\def\thefootnote{}
\begin{abstract}

Spoken language Identification (LID) systems are needed to identify the language(s) present in a given audio sample, and typically could be the first step in many speech processing related tasks such as automatic speech recognition (ASR). Automatic identification of the languages present in a speech signal is not only scientifically interesting, but also of practical importance in a multilingual country such as India. In many of the Indian cities, when people interact with each other, as many as three languages may get mixed. These may include the official language of that province, Hindi and English (at times the languages of the neighboring provinces may also get mixed during these interactions). This makes the spoken LID task extremely challenging in Indian context. While quite a few LID systems in the context of Indian languages have been implemented, most such systems have used small scale speech data collected internally within an organization. In the current work, we perform spoken LID on three Indian languages (Gujarati, Telugu, and Tamil) code-mixed with English. This task was organized by the Microsoft research team as a spoken LID challenge. In our work, we modify the usual spectral augmentation approach and propose a language mask that discriminates the language ID pairs, which leads to a noise robust spoken LID system. The proposed method gives a relative improvement of approximately 3-5\% in the LID accuracy over a baseline system proposed by Microsoft on the three language pairs for two shared tasks suggested in the challenge.

\end{abstract}

\section{Introduction}

In recent years, advances in artificial intelligence (AI)  have significantly expanded the degree to which individuals can interact with technology utilizing just their voice \cite{Intro_Speech_hinton2012deep}. However, these systems often require explicit information about the language of the users. The ability to dynamically process a single input speech stream consisting of different languages would expand the usefulness of existing voice-based systems and open up a wide array of additional functionalities. The spoken language identification (LID) research addresses this issue by exploring how to extract information from the audio signal and use it to predict the spoken language. LID is used in several applications such as multilingual translation systems or emergency call routing, where the response time of a fluent native operator might be critical \cite{LID_India_kumar2004language}.

Over the years, researchers have utilized many prosodic and acoustic features to construct machine learning models for LID systems \cite{LID_Propodic_HMM_obuchi2005language}. Several prosodic and acoustic features are based on phonemes, which become the underlying features that drive the performance of the statistical models \cite{LID_prosodic_safitri2016spoken}. If two languages have many overlapping phonemes, then identifying them becomes a challenging task for a classifier.  Subsequent LID systems relied on acoustic modelling \cite{LID_Acoustic_StModel_tong2006integrating}. In particular, guided by the advances on speaker verification, the use of i-vector feature extractors as a front-end followed by diverse classification mechanisms became popular as acoustic LID systems \cite{LID_AMSystem_gonzalez2010multilevel} \cite{LID_AMSystem_torres2010mitll}. The extensive feature engineering with i-vectors results in very complex systems, with an increasing number of computational steps in their pipeline \cite{LID_IVectors_martinez2011language}.

Approaches solely based on applying neural networks on input features like mel-frequency cepstral coefficients (MFCC) show that they reach state-of-the-art results, while being less complex \cite{LID_gelly2016language}. Current research on language identification systems using  deep neural networks (DNN) mainly focuses on using different forms of long short term memory (LSTMs), working on input sequences of transformed audio data.  The resulting sequence features are fused together and used to classify the language of the input samples.  In \cite{LID_ConvNet_shukla2019spoken} \cite{LID_CNN_bartz2017language}, a DNN based  architecture was proposed for extracting spatial features from the log-mel spectrograms of raw audio using convolutional neural networks (CNNs) and then using recurrent neural networks (RNNs) for capturing temporal features to identify the language.

Automatic spoken LID systems are particularly relevant for multilingual countries such as India. 
%More than 200 languages are spoken across the country. An LID system recognising 27 Indian languages was implemented using Gaussian Mixture Model (GMM) and MFCC features.  
An LID system recognising 27 Indian languages was implemented using Gaussian mixture model (GMM) and MFCC features \cite{LID_India_koolagudi2012identification}. Another LID system for Bengali, Hindi, Telugu, Urdu, Assamese, Punjabi and Manipuri languages was implemented that used feed forward neural networks trained with two hours of speech data of each of the seven languages \cite{LID_India_2_rao2015language}. Most LID systems for Indian languages were trained and tested with either speech of professional news readers \cite{LID_India_kumar2004language} or with small scale speech data collected by researchers \cite{LID_India_4_madhu2017automatic}  \cite{LID_India_5SLT_chakraborty2018language}, that may not be available for public use. Till recently, the lack of publicly available Indian language speech databases was a limitation for a scientific study of various acoustic analysis and modeling techniques. Recently, a phonetically balanced speech corpus of Hindi-English code-mixed speech was developed by IIT Guwahati \cite{CodeMix_Hindi_pandey2018phonetically}, and is explicitly designed for automatic speech recognition (ASR) paradigm, but not for the LID task. 
%Also, in most of the Indian cities, at least three languages are spoken, the official language of that province, Hindi, English and at times the languages of the neighboring provinces. Further, it is fairly common to mix words from English and Hindi along with the native language, and such code-mixing becomes challenging for spoken LID on Indian languages.

%Collecting speech data for Indian languages can be an expensive task, and alternative schemes are to be explored. 
Microsoft organized a shared task on Code-switched spoken language identification (LID) in three language pairs – Gujarati-English, Telugu-English and Tamil-English \cite{MicrosoftSpokenLID}. The shared task consists of two subtasks:(1) utterance-level identification of monolingual vs. code-switched utterances, and, (2) frame-level identification of languages in a code-switched utterance.

Spectral augmentation is a popular technique to improve the noise robustness of  an ASR system \cite{DataAug_ASR_ko2015audio} \cite{DataAug_ASR2_toth2018perceptually}. It involves three main steps namely time warping, frequency masking, and temporal masking. In this paper, we have explored the spectral augmentation \cite{SpecAug_park2019specaugment} for detecting the LID information. The spectral augmentation approach randomly chooses the position of the different masks. However, in our work we select the temporal masks based on the number and position of the English segments in a code-switched utterance. The clean and the masked spectrograms are provided as an input to the CNN-BiLSTM encoder network trained with CTC loss function.  The variation of frequency and temporal masks in the recent spectral augmentation \cite{SpecAug_park2019specaugment} are found to provide discriminative power to identify the language information in a code-mixed speech data.

The organization of the remaining sections of the paper are as follows: In Section 2, the importance of spectral augmentation for LID is discussed. The proposed framework for spoken LID is explained in Section 3. Section 4 provides the experimental set-up of the proposed spectral augmentation based spoken LID. We analyze the impact of the proposed approach on spoken LID performance in Section 5. Section 6 describes the conclusions and future directions.

\section{Spectral Augmentation}

In this section, the related work on spectral augmentation for different speech processing applications is discussed. Data augmentation is a popular method for improving robustness and training of neural networks \cite{DataAug_ASR_ko2015audio} . 
It's been used successfully in several domains starting from image classification to molecular modelling \cite{DataAug_bjerrum2017data}.
The fundamental principle is to increase the amount of training samples by creating multiple variants of the dataset. The data augmentation method is typically applied to create an additional training data for ASR to improve the performance. For instance, in \cite{DataAug_LowResource_ragni2014data}, the data was augmented for low resource speech recognition tasks. Vocal Tract Length Normalization has been explored for data augmentation in \cite{DataAug_VTLN_jaitly2013vocal}. Speech perturbation has been applied on raw audio for LVCSR tasks in \cite{DataAug_ASR2_toth2018perceptually}. The use of an acoustic room simulator has been adopted in \cite{DataAug_LowResource_ragni2014data}. Data augmentation is used to spot the important keywords from the speech utterance \cite{DataAug_KWS_raju2018data}. 
Perceptually, human listeners show remarkable tolerance to a variety of spectrotemporal manipulations of the input sound signal during segregation of speech-in-noise recognition tasks \cite{teki2011brain} \cite{teki2013segregation} \cite{teki2016neural}. Inspired by such studies in human auditory perception and cognitive neuroscience, and the recent success of augmentation techniques in the speech and vision domains, SpecAugment, an augmentation method that operates on the log mel spectrogram of the input audio (rather than the raw audio ) was proposed in \cite{SpecAug_park2019specaugment}. This method is simple and computationally inexpensive, as it directly acts on the log mel spectrogram as if it was an image.

Spectral Augmentation includes the following steps:
\begin{enumerate}
    \item \textbf{Time warping}:  Given a log mel spectrogram with $\tau$ time steps, it is viewed as an image where horizontal and vertical axes represents time (s) and the frequency (Hz) respectively. A random point along the horizontal line passing through the center of the image within the time steps ($W$, $\tau$ - $W$) is to be warped either to the left or right by a distance $w$ chosen from a uniform distribution from 0 to the time warp parameter $W$ along that line.
    \item \textbf{Frequency Masking}: It is applied so that $f$ consecutive mel frequency channels [$f_0$, $f_0$ + $f$) are masked, where $f$ is first chosen from a uniform distribution from 0 to the frequency mask parameter $F$, and $f_0$ is chosen from [0, $v$ - $f$). $v$ is the number of mel frequency channels.	
    \item \textbf{Temporal Masking}: It is applied so that $t$ consecutive time steps [$t_0$, $t_0$ + $t$) are masked, where $t$ is first chosen from a uniform distribution from 0 to the time mask parameter $T$, and $t_0$ is chosen from [0, $\tau$ - $t$).
\end{enumerate}

Figure 1 shows examples of the individual augmentations applied to a single speech input (PartBGujarati/Dev/Audio/000060438.wav). The log mel spectrograms are normalized to have zero mean value, and thus setting the masked value to zero is equivalent to setting it to the mean value.

\begin{figure}[h]
\begin{minipage}[b]{1.0\linewidth}
  \centering
  \centerline{\includegraphics[width=7.0cm,height=5.0cm]{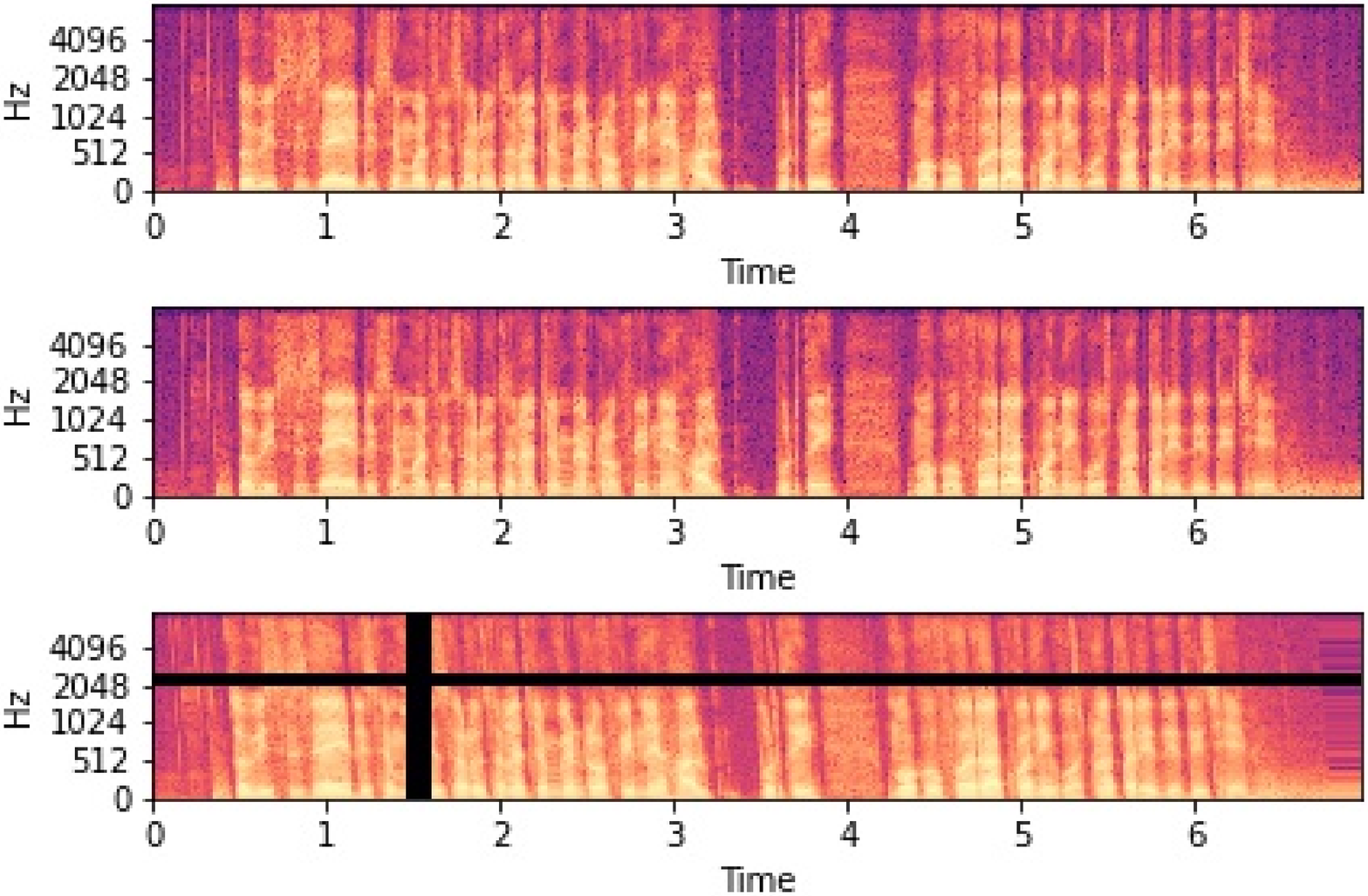}}
%  \vspace{2.0cm}
  %\centerline{(a) Result 1}\medskip
  \caption{Augmentations applied to the base input. The figures (from top) depict the log mel spectrogram of the base input with no augmentation, (time warp, frequency and time masking applied)
}
  \label{fig:SpecAug01}
\end{minipage}
\end{figure}

The spectral augmentation approach is shown to be successful on Noise robust ASR \cite{DataAug_ASR_ko2015audio}. For instance, the selection of different masks in the spectral augmentation can discriminate between the clean speech and the noisy speech. In  code-mixed speech spoken in India, it is generally observed that the native speaker tries to speak in her native language, and occasionally uses English words.  The process of masking the English words can lead to building a mono-lingual corpus. In order to build an efficient spoken LID system, the DNN may first learn to discriminate between the code-mixed speech and the mono-lingual speech. Since the temporal masking positions can be altered with respect to different languages, we sought to mask out the English words in the code-switched speech.

\section{Proposed Work}

Figure 2 shows the CNN-LSTM system that uses CTC loss function at the output layer. 
\begin{figure}[h]
\begin{minipage}[b]{1.0\linewidth}
  \centering
  \centerline{\includegraphics[width=7.5cm,height=2.5cm]{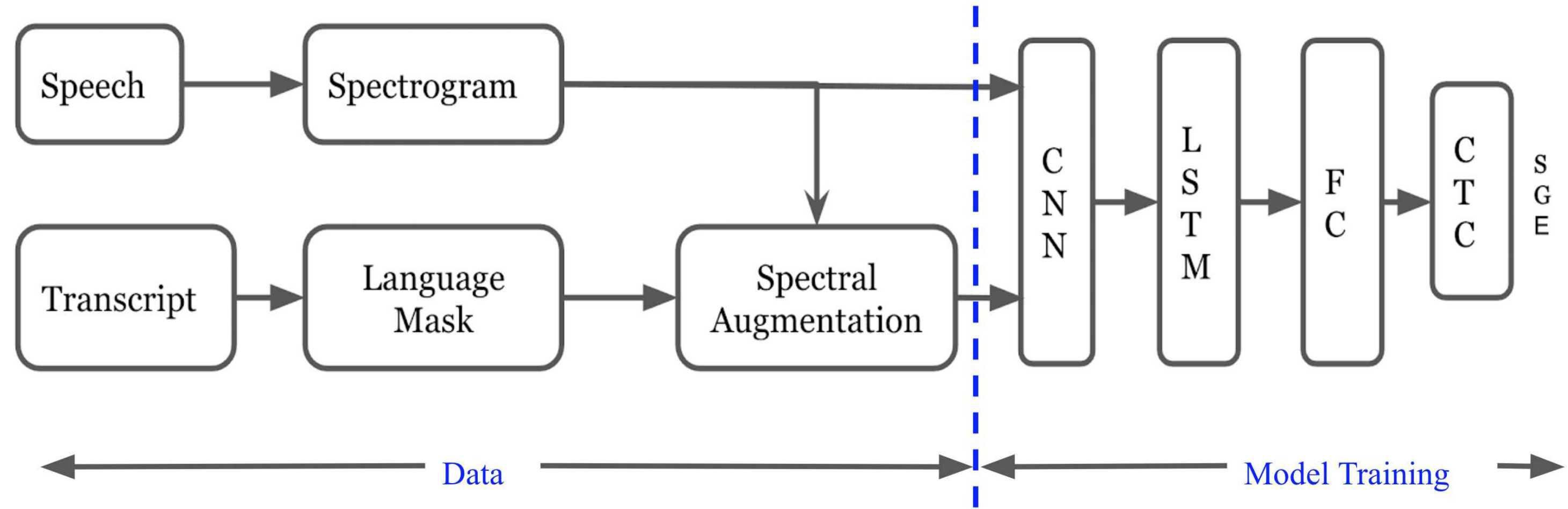}}
%  \vspace{2.0cm}
  %\centerline{(a) Result 1}\medskip
  \caption{End-to-End spoken LID using CNN-LSTM model }
  \label{fig:SpecAug01}
\end{minipage}
\end{figure}
The fundamental end-to-end CTC system receives the 2-D spectrogram as shown in the bottom of the figure. The CNN layer is effectively used in speech processing applications as they are capable of modeling temporal invariance for variable length utterances \cite{E2ECTC_miao2016empirical}. The BiLSTM models are trained on the convolved features to capture long-term sequential context. Later, the dense layer or the fully connected (FC) layer is connected to the output layer. The output is produced through a softmax function computing a probability distribution over the target labels. The target labels used in Figure 2 are {S, G and E} which corresponds to Silence, Gujarati and English language ID, respectively.  To accelerate the training procedure, Batch Normalization \cite{Intro_Speech_hinton2012deep} is applied on hidden layers. In our work, we feed the clean spectrogram, and the augmented spectrogram to the end-to-end LID system. However, the spectral augmentation is obtained by proposing an appropriate temporal mask termed as ‘language mask’ in the paper, and explained in the next subsection.

\subsection{Language Transcript based Temporal mask}
The conventional temporal mask(s) used in the spectral augmentation is selected at a random position that follows a uniform distribution from 0 to the time mask parameter $T$. In our work, at the time of training, the temporal masks are obtained from the language ID transcripts. The steps followed to obtain the temporal mask from the language transcript are as follows:

\begin{enumerate}
    \item Initially, the number of characters in the transcript is computed ($l_1$).
    \item Find the language segments from the transcript. Ex: If transcript = ‘SSSGGGEEGG’, then the language segments are {‘SSS’, ‘GGG’, ‘EE’, and ‘GG’}, where ‘S’, ‘E’ and ‘G’ corresponds to the silence, English, and Gujarati speech segments, respectively.
    \item Obtain the maximum number of frames based on the length of the characters in the transcript ($N_f$). Since the language label is created for every 200 ms for the shared task, every language ID character corresponds to 200 ms of speech content.
    \item We calculate the number of temporal masks by finding the number of speech segments corresponding to the English speech. Since, the non-predominant language in the code-mixed speech data is English, we mask the English words from the spectrogram.
    \item Obtain the start ($t_0$), and end time frame index ($t_1$) for every English speech segment. 
    \item The $t_0$ to $t_1$ consecutive time steps [$t_0$, $t_1$] are masked.
   
\end{enumerate}

Figure 3 shows a toy example to augment the spectrogram based on the language mask. The spectrogram is obtained for the speech signal ‘PartB/Gujarati/000010183.wav’ as shown in Figure 3 (a). The speech signal is sampled at 16 KHz with window size of 200 ms, and window stride of 100 ms. 
\begin{figure}[h]
\begin{minipage}[b]{1.0\linewidth}
  \centering
  \centerline{\includegraphics[width=7.5cm,height=5.0cm]{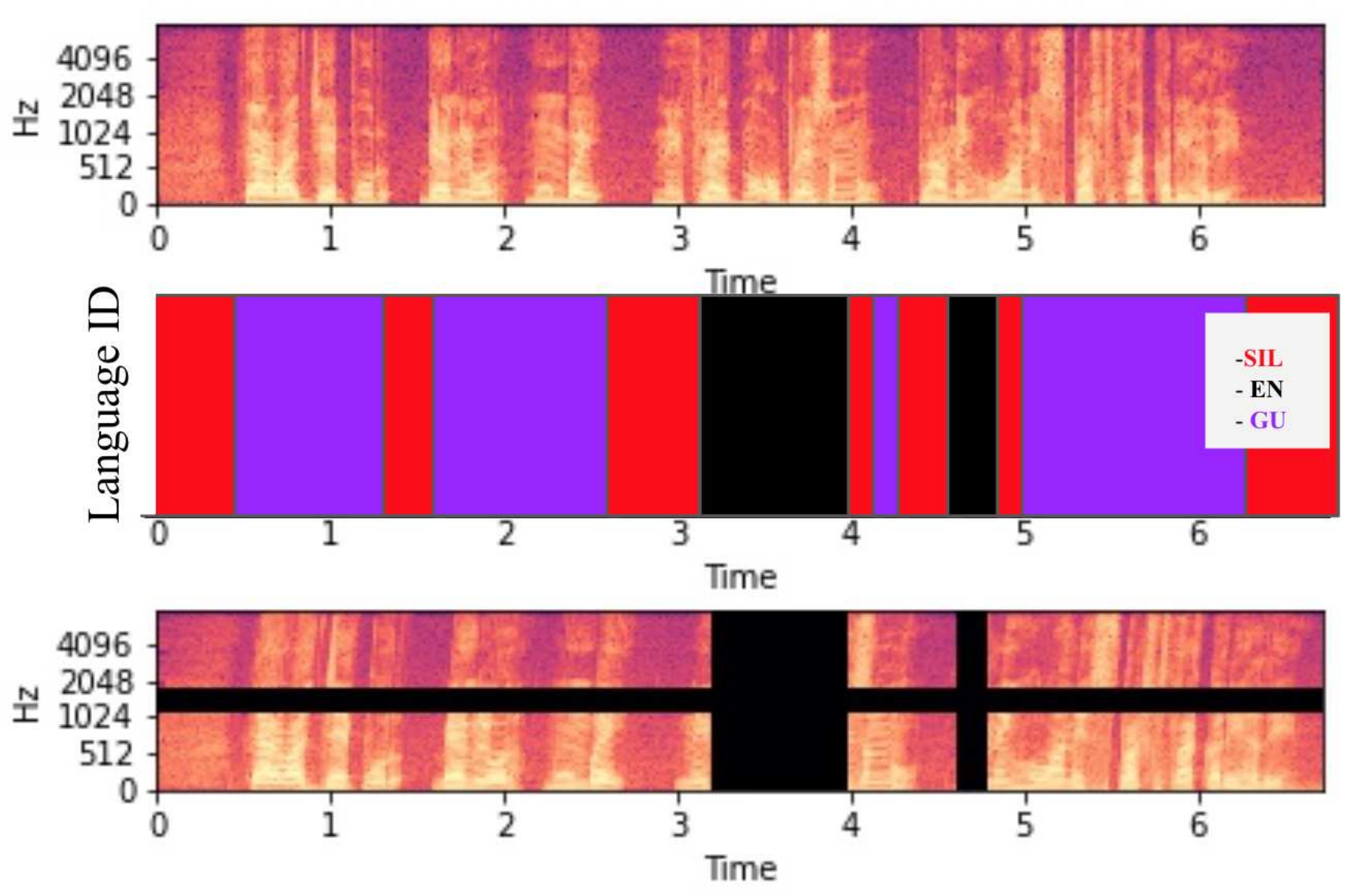}}
%  \vspace{2.0cm}
  %\centerline{(a) Result 1}\medskip
  \caption{Illustrative example of the proposed LID based temporal masking in spectral augmentation: (Top) Raw spectrogram, (Middle) Language information at different time instance, and (Bottom) Proposed spectral augmentation
}
  \label{fig:SpecAug01}
\end{minipage}
\end{figure}
The language transcripts are mapped to the speech frames, and are shown in Figure 3 (b). For instance, there is an English phrase (‘flight operation’) in the speech signal from 3.21 s to 3.99 s (shown in black). The number of temporal masks are decided based on the number of English segments present in the language transcripts. In this case, the number of English segments are two, and are present from (a) 3.1 - 3.99 s, and (b) 4.69 - 4.79 s. The corresponding time segments are masked as shown in Figure 3 (c) .

\section{Experiments}

First, we introduce our experimental environment and the metrics used for the spoken LID task. Then, we show our results on the shared tasks. Following this, we show the results of our experiments on the proposed work.

\subsection{Data}

The data set consists of three code-switched language pairs - Gujarati-English, Tamil-English and Telugu-English. 	The shared task for the workshop consists of two sub-tasks for spokenLID of code-switched audio. The two sub-tasks will consist of (1) Part A: Utterance-level LID, and (2) Part B: Frame-level LID .

% \usepackage{multirow}
\begin{table}[h]
\centering
\caption{Dataset description for the code-mixed spoken LID}
\footnotesize\setlength{\tabcolsep}{1.5pt}
\renewcommand{\arraystretch}{1.0}
\begin{tabular}{|c|c|c|c|c|}
\hline
\multirow{2}{*}{Data}   & \multirow{2}{*}{Language} & \multicolumn{3}{c|}{
Number of Samples (Duration in hh:mm:ss)
}   \\ \cline{3-5} 
                        &                           & Train            & Dev             & Test           \\ \hline
\multirow{3}{*}{Part-A} & Gu-En                     & 16780 (31:59:53) & 2091 (3:59:55)  & 2156 (3:44:50) \\ \cline{2-5} 
                        & Ta-En                     & 17943 (31:59:47) & 2258  (4:00:05) & 2258 (3:53:59) \\ \cline{2-5} 
                        & Te-En                     & 16991 (31:59:50) & 2135 (4:00:06)  & 2064 (3:42:15) \\ \hline
\multirow{3}{*}{Part-B} & Gu-En                     & 8620 (15:59:56)  & 1080(1:59:58)   & 1078 (2:00:02) \\ \cline{2-5} 
                        & Ta-En                     & 8982 (15:59:57)  & 1135 (1:59:59)  & 1129 (1:59:57) \\ \cline{2-5} 
                        & Te-En                     & 8226 (15:59:56)  & 1047 (2:00:03)  & 1033 (1:59:55) \\ \hline
\end{tabular}
\end{table}

\subsection{Training}

%The training phase is based on the open source DeepSpeech network [24], trained with the CTC activation function. 
The Microsoft's baseline system  is made up of an end-to-end multi-layer model consistingof 5 layers of LSTM, each consisting of 1024 neurons. The model is based on deepspeech-2 \cite{amodei2015deep}. The model is trained using the CTC loss function. 
The LID detector using CTC starts with two layers of 2D convolutions over both time and frequency domains with 32 channels, 41 × 11, 21 × 11 filter dimensions, and 2 × 2, 2 × 1 stride. Next, five Bi-LSTM layers with 1024 hidden units are followed by one fully connected linear layer with 5 softmax outputs {blank, $′$ , Gujarati, English, Silence}. The Bi-LSTM models have around 10.2 millions (M) parameters. The input sequence are values of spectrogram slices, 20 ms long, computed from Hamming windows with 10 ms frame shifts. The output (target) sequence was obtained directly from the letters of the word transcription. We used 100 epochs to train all the models used for further evaluation. The two variants of spectral augmentation are implemented in this paper: (1) spectral augmentation with random temporal, and frequency mask positions, and (2) spectral augmentation with the proposed language mask.

\subsection{Decoding}

Assuming an input sequence of length $T$ , the output of the neural network will be $p(c| x_t )$ for $t = 1, . . . , T$ . Let $p(c|x_t)$ is a distribution over possible characters in the alphabet $\psi$ (which includes the blank symbol) given audio input $x_t$. In order to recover a character string from the output of the neural network, as a first approximation, we take the argmax at each time step. Let $S = (s_1 , . . . , s_T )$ be the character sequence where 
%$s_t = \underset{c\in  \psi}{\arg\max}\ p(c|x_t )$
\begin{eqnarray}
s_t = \underset{c\in  \psi}{\arg\max}\ p(c|x_t )
\end{eqnarray}

%\begin{eqnarray}
%s_t = arg max c∈Σ p(c; x_t )
%\end{eqnarray}

 The sequence $S$ is mapped to a transcription by collapsing repeat characters and removing blanks. On the other hand, the beam search decoder uses the context information in generating the decoded sequence. There are two cases: either we extend the beam by a character $c$ different from the last character, then there is no need for separating blanks in the paths, or the last character is repeated.

\subsection{Evaluation}

For task A, the predicted label (Monolingual or code- switched) file is submitted for a particular audio file in the blind test set. For task B, a frame-level (200ms) label for each frame in the audio is submitted. The LID system performance is evaluated using accuracy and Equal Error Rate (EER) as evaluation metrics. 

\begin{equation}
Accuracy = \frac{N}{T}
\end{equation}
where $N$ and $T$ are the total no. of correctly predicted data samples and the total no. of data points in the speech dataset respectively.

\begin{equation}
EER = \frac{FRR+FAR}{2}
\end{equation}

where $FRR = {TFR}/{T} $, and $FAR = {TFA}/{T} $; $FRR$, $FAR$, $TFR$, $TFA$ and $T$ corresponds to false rejection rate, false acceptance rate, total no. of false rejects, total no. of false accepts and total number of datapoints, respectively.

%\subsection{Illustrative example - Baseline LID vs Proposed LID}

\section{Results and Discussion}

The Deepspeech-2 model is trained on the baseline configurations and on the variants of spectral augmentation. In this section, we report the accuracies and EERs on the Test set for Task-A and Task-B.  

Table \ref{tab:PartAResults} shows the \% spoken LID Accuracy of the baseline model and the proposed spectral augmentation method. It can be observed that the proposed system outperforms as compared to that of the baseline model on Task-A.		
\begin{table}[h]
\centering
\footnotesize\setlength{\tabcolsep}{2.5pt}
\renewcommand{\arraystretch}{1.0}
\caption{Spoken LID accuracy on Task-A for different datasets}
\label{tab:PartAResults}
\arrayrulecolor{black}
%\begin{tabular}{!{\color[rgb]{0.62,0.62,0.62}\vrule}l!{\color{black}\vrule}l!{\color{black}\vrule}l!{\color{black}\vrule}} 
\begin{tabular}{|c|c|c|}
\hline
Data\_Test & \% Acc [Baseline] & \% Acc [SpecAug]  \\ 
\hline
Gu-En      & 71.9                        & 73.01                        \\ 
\hline
Ta-En      & 71.2                        & 79.02                        \\ 
\hline
Te-En      & 74.0                        & 78.65                        \\
\hline
\end{tabular}
\arrayrulecolor{black}
\end{table}

For Task-B, we compare the performance of the code-mixed language recognizer using spectral augmentation with random temporal mask location, and the proposed language mask. Table \ref{tab:RES_PartB_GREEDYSEARCH} shows the performance comparison of the proposed work, and the DeepSpeech-2 model (Microsoft's Baseline) on the test set of Part-B with Greedy Search Decoding. It can be observed that the proposed language mask on the spectrogram is able to discriminate the language ID’s within an utterance.

% \usepackage{multirow}
\begin{table}[]
\centering
\footnotesize\setlength{\tabcolsep}{1.5pt}
\renewcommand{\arraystretch}{1.0}
\caption{Performance comparison of the proposed work, and the DeepSpeech-2 model (Microsoft's Baseline) on the test set of Part-B by Greedy Search Decoder
}
\label{tab:RES_PartB_GREEDYSEARCH}
\begin{tabular}{|c|c|c|c|}
\hline
\multirow{2}{*}{Language} & \multicolumn{3}{c|}{\% Acc {[}\%EER{]} on the test set on Greedy search decoder} \\ \cline{2-4} 
                          & Baseline                 & SpecAug                  & SpecAug + Lang Mask        \\ \hline
Gujarati                  & 66.79 {[}9.71{]}         & 75.33 {[}7.78{]}         &       
75.72 [7.53]
                    \\ \hline
Tamil                     & 72.17 {[}8.42{]}         & 74.80 {[}7.76{]}         & 75.02 {[}7.67{]}           \\ \hline
Telugu                    & 70.54 {[}8.65{]}         & 74.06 {[}7.88{]}         & 74.08 {[}7.87{]}           \\ \hline
\end{tabular}
\end{table}

Table \ref{tab:RES_PartB_BEAMSEARCH} shows the performance comparison of the proposed work, and the state-of-the-art methods on the test set of Part-B on Beam Search Decoder. The size of the beam width is varied from 5 to 20 in steps of 5. It is observed that the beam width size of 15 is optimal for all languages in the shared task.

% \usepackage{multirow}
\begin{table}[]
\centering
\footnotesize\setlength{\tabcolsep}{1.5pt}
\renewcommand{\arraystretch}{1.0}
\caption{Performance comparison of the proposed work, and the DeepSpeech-2 model (Microsoft's Baseline) on the test set of Part-B by Beam Search Decoder
}
\label{tab:RES_PartB_BEAMSEARCH}
\begin{tabular}{|c|c|c|c|}
\hline
\multirow{2}{*}{Language} & \multicolumn{3}{c|}{\% Acc {[}\%EER{]} on the test set on Beam search decoder} \\ \cline{2-4} 
                          & Baseline                & SpecAug                 & SpecAug + Lang Mask        \\ \hline
Gujarati                  & 72.53 {[}8.54{]}        & 76.33 {[}7.52{]}        &            76.64 [7.36]                \\ \hline
Tamil                     & 73.89 {[}8.06{]}        & 75.80 {[}7.54{]}        & 76.06 {[}7.44{]}           \\ \hline
Telugu                    & 75.20 {[}7.68{]}        & 75.90 {[}7.47{]}        & 75.84 {[}7.47{]}           \\ \hline
\end{tabular}
\end{table}

\subsection{Discussion}

The research on spoken LID for code-switched speech is relatively new for Indian languages. Further, in India it is fairly common to mix words from English and Hindi along with the native language. Spectral augmentation has shown promising improvement on noise robust ASR, and on low resource languages. Instead of masking random positions in the spectrogram, the positions of the language transitions are masked. Due to this, the train speech corpus contains only the code- switched, and the mono-lingual utterances. As a result, the model learns to discriminate well between the code-switched utterances, and the mono-lingual utterances. Also, the positions of the proposed language mask captures the corresponding grammar in the utterance. Thereby the model learns about the different statistics of code-mixing. On the speech segments where there are small portions of non-dominant language is involved, the model captures these small language transitions.

\section{Conclusion}

Spoken LID is performed on the three Indian languages (Gujarati, Telugu, and Tamil) code-switched with English. We proposed a language mask from the speech transcripts, and incorporated it in the spectrogram. We observed that the proposed method is able to discriminates the languages in a code-mixed data. The proposed method gives a relative improvement in performance of approximately 3-5\% in the LID accuracy over that of a baseline system proposed by Microsoft on three code-switched language pairs for two different  shared tasks.	 	

\balance

\bibliographystyle{IEEEtran}

%\bibliography{mybib}

% \begin{thebibliography}{9}
% \bibitem[1]{Davis80-COP}
%   S.\ B.\ Davis and P.\ Mermelstein,
%   ``Comparison of parametric representation for monosyllabic word recognition in continuously spoken sentences,''
%   \textit{IEEE Transactions on Acoustics, Speech and Signal Processing}, vol.~28, no.~4, pp.~357--366, 1980.
% \bibitem[2]{Rabiner89-ATO}
%   L.\ R.\ Rabiner,
%   ``A tutorial on hidden Markov models and selected applications in speech recognition,''
%   \textit{Proceedings of the IEEE}, vol.~77, no.~2, pp.~257-286, 1989.
% \bibitem[3]{Hastie09-TEO}
%   T.\ Hastie, R.\ Tibshirani, and J.\ Friedman,
%   \textit{The Elements of Statistical Learning -- Data Mining, Inference, and Prediction}.
%   New York: Springer, 2009.
% \bibitem[4]{YourName17-XXX}
%   F.\ Lastname1, F.\ Lastname2, and F.\ Lastname3,
%   ``Title of your INTERSPEECH 2020 publication,''
%   in \textit{Interspeech 2020 -- 20\textsuperscript{th} Annual Conference of the International Speech Communication Association, September 15-19, Graz, Austria, Proceedings, Proceedings}, 2020, pp.~100--104.
% \end{thebibliography}

\end{document}